\documentclass[letterpaper, 10 pt, conference]{ieeeconf}

\ifdefined\pdfsuppressptexinfo
\fi

\IEEEoverridecommandlockouts

\usepackage{color}
\usepackage{microtype}
\usepackage{makecell}
\usepackage{multirow}
\usepackage{booktabs}
\usepackage{url}
\makeatletter
\def\endfigure{\end@float}
\def\endtable{\end@float}
\makeatother
\usepackage[caption=false,font=footnotesize]{subfig}
\usepackage{threeparttable}
\usepackage{cite}

\usepackage{float} 
\usepackage{stfloats}
\usepackage{amssymb}
\usepackage{graphicx}
\usepackage{amsmath}
\usepackage{algorithm}
\usepackage{bm}
\usepackage{algpseudocode}
\usepackage{placeins}
\usepackage{siunitx}
\usepackage{tabularx}
\usepackage{array}
\usepackage[hidelinks]{hyperref}
              
\title{State-Dependent Objective Coordination for End-to-End Balance and Navigation}
\author{
XuHao Huang,
Xitao Xu,
Yuxin Zheng,
Yunfan Ren,
Zheng Gong*
}

\begin{document}

\setlength{\textfloatsep}{8pt plus 2pt minus 2pt}
\setlength{\floatsep}{6pt plus 2pt minus 2pt}
\setlength{\intextsep}{8pt plus 2pt minus 2pt}
\setlength{\dbltextfloatsep}{8pt plus 2pt minus 2pt}
\setlength{\dblfloatsep}{6pt plus 2pt minus 2pt}

\maketitle
\thispagestyle{empty}
\pagestyle{empty}

\begin{abstract}
Robotic tasks with coupled control objectives require trade-offs that vary with the system state. On two-wheeled inverted-pendulum robots, balance recovery and obstacle-avoiding navigation share the same wheel-torque inputs, creating competing control demands. We propose State-Dependent Objective Coordination (SDOC), a lightweight reward-level mechanism for training a unified end-to-end policy. An analytical mapping from pitch magnitude adjusts the relative weights of balance and navigation rewards, emphasizing navigation near the upright state and increasing the balance weight as attitude deviation grows. This coefficient guides the training of a unified end-to-end policy that jointly optimizes balance and navigation, whose control demands may conflict depending on the robot state. The deployed policy directly maps onboard observations to wheel torques. Gradient analysis at fixed policy checkpoints reveals attitude-dependent directional competition between the objectives. The policy is trained in NVIDIA Isaac Sim using parallel robot instances evenly distributed across randomly generated obstacle maps at multiple difficulty levels. Simulation training accounts for actuator discrepancies and asynchronous multimodal sensor observations with sensor-specific noise, and randomizes the robot's inertial parameters to improve policy robustness. Comparative experiments show that the policy trained with the proposed method achieves high navigation success rates while maintaining small attitude excursions across different obstacle layouts. Hardware experiments demonstrate zero-shot transfer without policy adaptation, with the policy maintaining balance and performing goal-directed navigation and obstacle avoidance under disturbances and in the presence of dynamic obstacles. Training and evaluation code will be made available in an anonymous repository at \url{XXX}.
\end{abstract}

\section{INTRODUCTION}

Real-world robot deployment increasingly requires compact control policies that satisfy multiple objectives within a single dynamical system under limited onboard computation and strict latency constraints. In strongly coupled robotic systems, these objectives may induce state-dependent conflicts, where an action that improves one objective can degrade another. Existing studies coordinate locomotion and manipulation through modular or hierarchical architectures~\cite{huang2024hilma,ma2022combining}, whereas task progress and sustained stability require state-dependent coordination when they share the same control inputs.


\begin{figure}[!t]
\centering
\includegraphics[width=0.9\columnwidth]{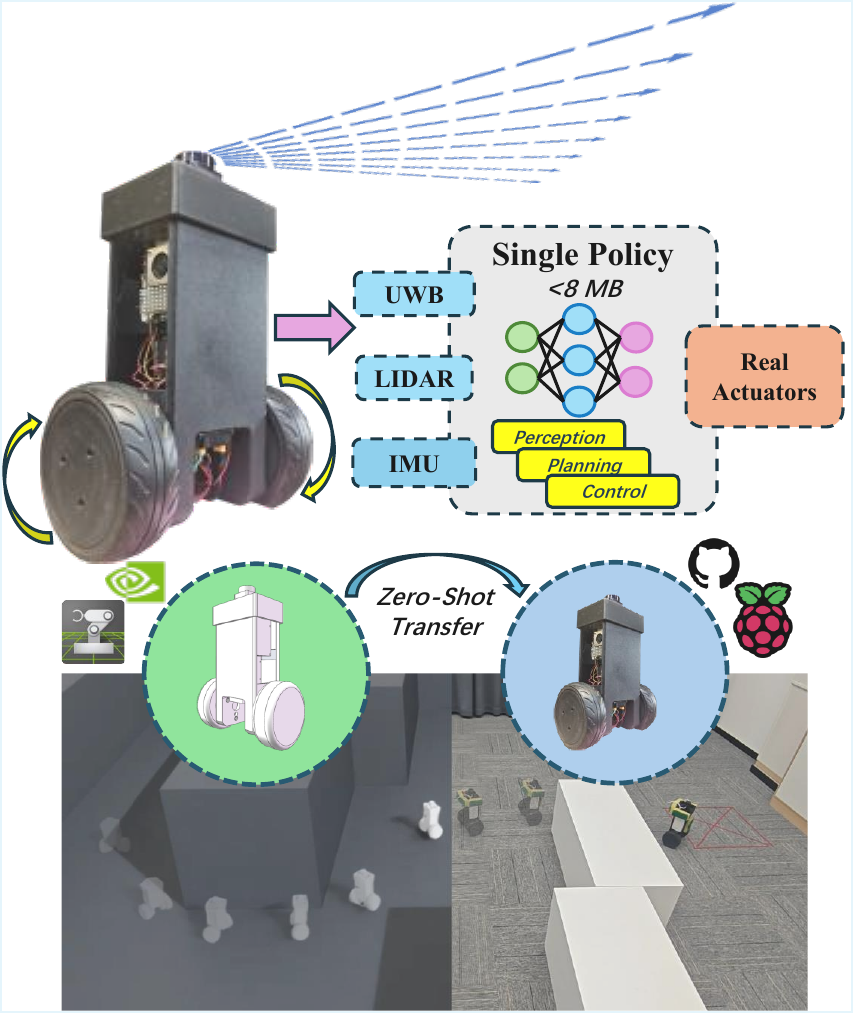}
\caption{Zero-shot simulation-to-real-world transfer of the learned policy using ultra-wideband (UWB), light detection and ranging (LiDAR), and an inertial measurement unit (IMU). Left: simulation. Right: real-world deployment.}
\label{fig:teaser}
\end{figure}

We study this problem on a two-wheeled inverted-pendulum robot that must maintain attitude stability while performing goal-directed obstacle-avoiding navigation. Navigation progress, obstacle avoidance, and attitude stabilization all act through the same wheel-torque commands. Near the upright state, excessive emphasis on stabilization may suppress effective navigation. At larger attitude deviations, aggressive navigation and avoidance actions may reduce recoverability and increase the risk of falling. Therefore, the appropriate balance between stabilization and navigation varies with the current attitude state.


A fixed reward coefficient applies the same objective weighting across states and therefore cannot explicitly adapt to state-dependent attitude risk. A large stabilization weight may be overly conservative near the upright state, whereas a small weight may provide insufficient stabilization as attitude risk increases. Structural alternatives, such as stage-wise training, hierarchical control, and policy composition, do not necessarily adjust this trade-off according to the current attitude state.


To address this issue, we propose State-Dependent Objective Coordination (SDOC), a lightweight reward-level coordination mechanism that analytically determines a scalarization coefficient $\beta(s)\in[0,1]$ from the robot's attitude state. During training, $\beta(s)$ adjusts the relative contributions of balance stabilization and navigation progress according to the robot's pitch magnitude. Navigation receives greater relative emphasis near the upright state, while stabilization is progressively emphasized as the pitch deviation grows. The coefficient is used only for reward construction during training. At deployment, a single end-to-end actor maps onboard observations directly to wheel torques without online reward computation, policy switching, or an auxiliary stabilizing controller.


We further incorporate deployment-relevant discrepancies into simulation training, including execution delay, action hold, asynchronous and noisy observations, and dynamics uncertainty. The resulting actor is trained in NVIDIA Isaac Sim and directly transferred to the physical robot without policy adaptation, as illustrated in Fig.~\ref{fig:teaser}. Hardware experiments demonstrate navigation in unseen environments and under disturbances.


\noindent\textbf{Our contributions are as follows:}
\begin{itemize}
\item We formulate simultaneous balance and obstacle-avoiding navigation as a state-dependent objective coordination problem and propose SDOC, a lightweight reward-level coordination mechanism that analytically determines a scalarization coefficient $\beta(s)$ from pitch for training a unified end-to-end policy.
\item We use gradient analysis at fixed policy checkpoints to examine attitude-dependent directional competition between the balance and navigation objectives, and compare the proposed method with alternative reward-weighting, constraint, training, and control-structure strategies.
\item We demonstrate zero-shot policy deployment on a physical two-wheeled inverted-pendulum robot, validating balance and obstacle-avoiding navigation in unseen environments, under external disturbances, and in the presence of dynamic obstacles.
\end{itemize}

\section{RELATED WORK}

Strongly coupled robotic control tasks cannot always be decomposed into independent subproblems. To coordinate different control demands, existing studies organize complex behaviors through skill decomposition, policy switching, and hierarchical control~\cite{dao2024sim,he2024learning,lee2024learning}, and address posture and motion tasks through posture manipulation, trajectory planning, MPC, or learning-based balance control~\cite{salagame2024heading,cui2021learning,DBLP:conf/icra/WangXFMXCLG23,krupa2021real}. In two-wheeled inverted-pendulum robots, maintaining balance is a prerequisite for continued navigation, and the relative priorities of balance and navigation vary with robot attitude. When a single end-to-end policy handles balance and navigation demands, the learning process needs to account for this state-dependent objective trade-off.


Reward weighting and scalarization are widely used for multi-objective coordination in reinforcement learning, ranging from fixed weighted combinations~\cite{van2014multi} to dynamic weighting, preference adaptation, and auxiliary rewards that adjust objective priorities with training or task requirements~\cite{abels2019dynamic,buet2023robust,wang2025reward}. Recent robotics studies further incorporate preferences, constraints, and task stages: MORLAX uses preference-conditioned hypernetworks to learn a family of policies representing different trade-offs~\cite{11551332}; AMOR learns a reward-weight-conditioned motion-tracking policy and further uses a high-level policy to select weights dynamically from the current state and motion context~\cite{alegre2025amor}; ROGER adjusts reward gains based on constraint-related penalties to coordinate task performance and constraint satisfaction~\cite{SrisuchinnawongA-RSS-25}; and CoMOPPO combines reward and cost design for predefined task stages to coordinate multiple objectives and constraints~\cite{kim2025stage}.


Beyond coordination through rewards and preferences, another line of research addresses gradient conflicts among multiple objectives during policy optimization. ESPO adaptively adjusts the number of samples used for policy updates based on gradient conflicts between reward and safety objectives~\cite{gu2024enhancing}; CoMOGA reformulates multi-objective optimization as a constrained optimization problem and coordinates conflicting objectives through constrained policy updates~\cite{kim2025conflict}; and GCR-PPO combines per-objective value estimation with gradient conflict handling to coordinate multiple objectives in robot reinforcement learning~\cite{munn2026scalable}.


These studies focus on preference-conditioned policy generation, learned weight selection, constraint-feedback-driven reward gain adjustment, stage-wise reward design, or gradient conflict handling. This work considers robotic platforms that require continuous active balancing, where balance is a prerequisite for continued navigation and attitude recovery and navigation progress can impose conflicting demands on the same control inputs. To address this coupling, SDOC coordinates balance and navigation through lightweight attitude-conditioned reward weighting. The scalarization coefficient is determined analytically and directly from the robot's physical state, incorporating attitude-dependent objective priorities into reward construction to guide the joint optimization of balance and obstacle-avoiding navigation within a single end-to-end policy. Task progress receives greater relative emphasis near upright, while balance recovery is emphasized at larger attitude deviations. The trained policy directly maps onboard observations to wheel torques for deployment.


\begin{figure*}[!t]
    \centering
    \includegraphics[width=0.85\linewidth]{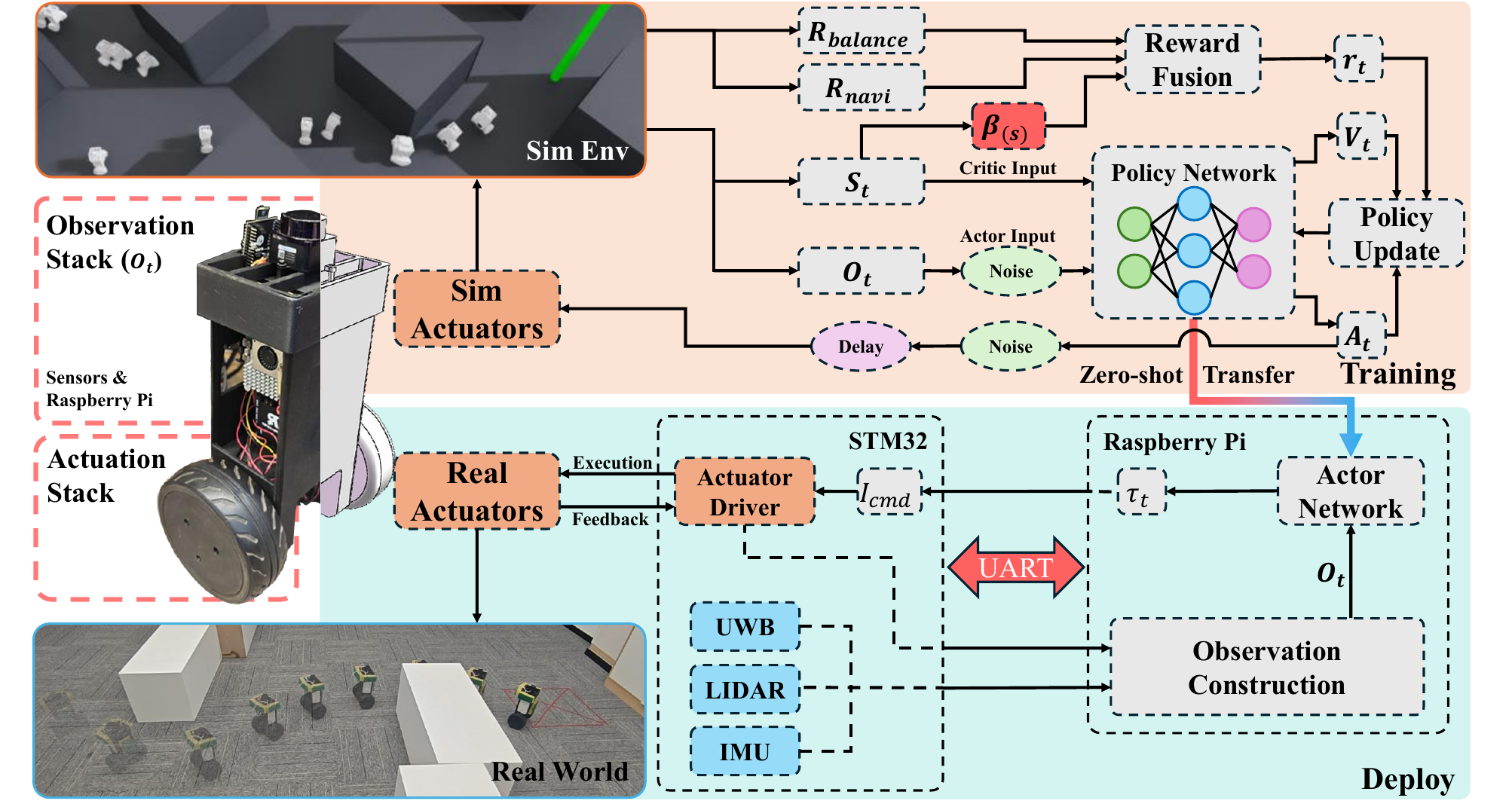}
\caption{
SDOC and deployment-oriented training. During simulation training, $\beta(s_t)$ performs reward-level state-conditioned scalarization of balance and navigation, while safety regularization is separated. Training accounts for actuator discrepancies, asynchronous multimodal observations with sensor-specific noise, and domain randomization of inertial parameters. The trained actor is extracted from the policy network and deployed using onboard observations only.
}
    \label{fig:system_framework}
\end{figure*}

\section{PROBLEM FORMULATION}
\label{sec:problem_formulation}

\subsection{Task Formulation}
\label{subsec:task_formulation}

A single end-to-end policy maps observations to left- and right-wheel torques, enabling the two-wheeled inverted-pendulum robot to reach the goal while maintaining balance and avoiding obstacles. We formulate the task as a partially observable Markov decision process (POMDP):
\begin{equation}
\mathcal{M}=\left\langle\mathcal{S},\mathcal{A},\mathcal{P},\mathcal{O},r,\gamma\right\rangle,
\label{eq:pomdp_formulation}
\end{equation}
where $\mathcal{S}$ and $\mathcal{A}$ denote the underlying state and action spaces, $\mathcal{P}$ represents the transition dynamics, $\mathcal{O}$ denotes the observation space, $r$ is the per-step reward, and $\gamma$ is the discount factor. The policy maps observation history to wheel torques. Episodes end upon goal reaching, falling, exceeding the collision limit, or timeout.

The observation contains attitude-related proprioception, goal-relative navigation information, and obstacle observations:
\begin{equation}
\begin{aligned}
o_t&=\left[o_t^{\mathit{att}},\;o_t^{\mathit{obs}},\;o_t^{\mathit{goal}}\right]\in\mathbb{R}^{27},\\
o_t^{\mathit{att}}&=\left[\theta_t,\;\dot{\theta}_t,\;\dot{\psi}_t,\;\omega_{L,t},\;\omega_{R,t}\right],\\
o_t^{\mathit{obs}}&=\left[\ell_{t,1},\ldots,\ell_{t,20}\right],\qquad
o_t^{\mathit{goal}}=\left[d_t,\;c_t\right].
\end{aligned}
\label{eq:state_definition}
\end{equation}
Here, $o_t^{\mathit{att}}$ contains proprioceptive information about body attitude and wheel motion: $\theta_t$ is the IMU-estimated pitch; $\dot{\theta}_t$ and $\dot{\psi}_t$ are the body-frame angular velocities about the pitch and yaw axes provided by the IMU; and $\omega_{L,t}$ and $\omega_{R,t}$ are the left and right wheel angular velocities obtained from motor feedback.

The LiDAR is mounted on top of the robot body, and its ranging rays tilt with body pitch in both simulation and hardware, without additional horizontal compensation. Twenty ranging rays are uniformly arranged across the frontal $180^\circ$ field of view to sense obstacle distribution. The ranges are clipped to $[0,2]$ m and arranged in a fixed ray order to form the obstacle observation $o_t^{\mathit{obs}}$ fed into the policy network.

Robot position and attitude are read directly in simulation. On hardware, three fixed UWB anchors and one robot-mounted tag estimate position, while the IMU attitude quaternion determines heading. The robot position, heading, and predefined goal position determine the goal distance $d_t$ and the cosine $c_t$ of the angle between the goal direction and robot heading. Together, they form the goal-relative observation $o_t^{\mathit{goal}}$, providing relative distance and direction information.

Simulation models observation noise separately for each sensor modality to account for their different noise characteristics, and applies domain randomization to the robot's inertial parameters to improve robustness. Consistent observation definitions across simulation and hardware support direct transfer and deployment of the trained policy. At deployment, the policy uses real-time obstacle observations and goal-relative information without a prebuilt obstacle map.

\subsection{State-Dependent Objective Conflict}
\label{subsec:objective_conflict}

During policy optimization, balance and navigation may favor competing update directions. To characterize this competition, we freeze the policy parameters and use the same sampled data to compute each objective's policy gradient, identifying directional conflict by a negative inner product. Gradient analysis of policy models from different stages of the same training run shows that directional competition varies with robot attitude, providing empirical motivation for state-dependent objective trade-offs.

Physical task requirements determine the appropriate trade-off across attitudes. Near upright with small attitude variations, the relative balance weight should be reduced to prevent excessive stabilization from restricting navigation. As attitude deviation grows, balance deserves higher priority as a prerequisite for continued navigation; its relative weight should therefore increase to promote attitude recovery.

Fixed reward weights cannot readily represent this attitude-dependent trade-off. Dynamic weights based on training progress or aggregate optimization statistics can adjust objective trade-offs, but may not capture immediate priority changes between task progress and attitude recovery unless they account for the current robot state. We therefore propose SDOC to dynamically adjust the relative weights of balance and navigation according to robot state, aligning objective coordination with current physical task requirements.

\section{METHODOLOGY}
SDOC coordinates balance stabilization and navigation progress through state-dependent reward weighting. The coefficient $\beta(s_t)$ is determined by the current robot attitude and adjusts the relative weights of the balance and navigation rewards. Fig.~\ref{fig:system_framework} summarizes the complete pipeline, and the following subsections specify the scalarization coefficient and reward components.

\subsection{State-Dependent Objective Coordination}

We determine a coefficient $\beta(s_t)\in[0,1]$ from the current robot state to weight the balance and navigation rewards. The per-step reward is defined as

\begin{equation}
r_t=\beta(s_t)\,r_{\mathit{bal}}+\bigl(1-\beta(s_t)\bigr)\,r_{\mathit{nav}}+r_{\mathit{safe}},
\label{eq:method_reward_decomp}
\end{equation}

where $r_{\mathit{bal}}$, $r_{\mathit{nav}}$, and $r_{\mathit{safe}}$ denote the balance-stabilization, navigation-progress, and safety-regularization terms, respectively. The safety term is added independently. Unlike fixed or globally adapted coefficients, $\beta(s_t)$ explicitly adjusts the relative objective preference across attitude states.

\begin{figure}[h]
  \centering
  \includegraphics[width=\linewidth]{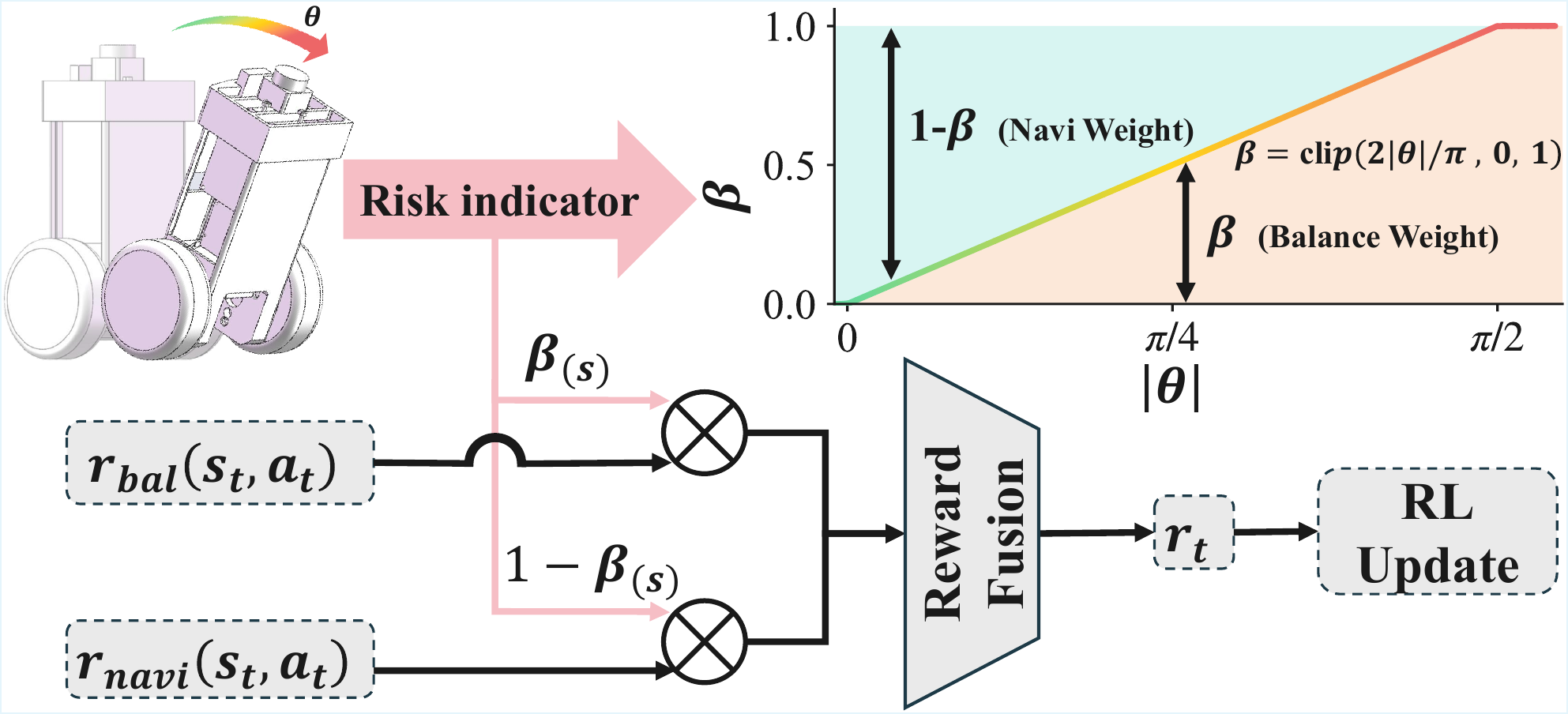}
  \caption{SDOC on the balancebot. A pitch-based attitude-state indicator determines the scalarization coefficient $\beta(s_t)$, which adjusts the relative contributions of the balance-stabilization and navigation-progress objectives. The coefficient increases with pitch magnitude according to Eq.~\eqref{eq:beta_inst}.}
  \label{fig:beta_intuition}
\end{figure}

For the two-wheeled inverted-pendulum platform, pitch magnitude measures deviation from upright. We therefore use a linear weighting map based on pitch magnitude to incorporate attitude information and the corresponding objective priorities into reward construction:

\begin{equation}
\beta(s_t)=\mathrm{clip}\left(\frac{2|\theta_t|}{\pi},0,1\right),
\label{eq:beta_inst}
\end{equation}

where $\theta_t$ is the pitch angle in radians. The function $\mathrm{clip}(\cdot,0,1)$ bounds the weight within $[0,1]$. Fig.~\ref{fig:beta_intuition} illustrates the mapping. During training, $\beta(s_t)$ is computed from the current attitude at each time step to construct the weighted reward, guiding policy learning through returns and advantage estimates.

\subsection{Reward Design}

Following the reward decomposition in Eq.~\eqref{eq:method_reward_decomp}, the balance-stabilization reward encourages an upright posture and attitude stability:

\begin{equation}
r_{\mathit{bal}}=1-c_{\theta}\mathcal{P}_{\theta}(|\theta_t|)-c_{\dot{\theta}}|\dot{\theta}_t|-c_{\dot{\psi}}\dot{\psi}_t^2,
\label{eq:r_bal}
\end{equation}

where $\theta_t$, $\dot{\theta}_t$, and $\dot{\psi}_t$ denote the pitch angle, pitch angular velocity, and yaw angular velocity, respectively. The pitch-risk penalty is defined as

\begingroup
\small
\begin{equation}
\mathcal{P}_{\theta}(|\theta_t|)
=
\begin{cases}
|\theta_t|^2,
& |\theta_t| \leq \theta_{\mathit{safe}}, \\[1mm]
|\theta_t|^2
+
\lambda_{\theta}
\bigl(|\theta_t|-\theta_{\mathit{safe}}\bigr)^2,
& |\theta_t| > \theta_{\mathit{safe}} .
\end{cases}
\label{eq:p_theta_definition}
\end{equation}
\endgroup

Here, $\theta_{\mathit{safe}}$ is the threshold for activating the additional penalty, and $\lambda_{\theta}$ controls its strength. Above this threshold, the additional term strengthens the penalty on larger attitude deviations.

The navigation-progress term is

\begin{equation}
\begin{aligned}
r_{\mathit{nav}}
&=
c_d \Delta d_t
+
c_{\psi}\cos(\Delta\psi_t)
+
c_{\mathit{goal}}\mathbb{I}_{\mathit{goal}} \\
&\quad
-
c_{\mathit{obs}}\mathcal{P}_{\mathit{obs}}(d_{\mathit{obs}})
-
c_{\mathit{col}}\mathbb{I}_{\mathit{col}}
-
c_a \|a_t-a_{t-1}\|^2,
\end{aligned}
\label{eq:r_nav}
\end{equation}

where $\Delta d_t$ denotes the reduction in distance to the target, $\Delta\psi_t$ is the goal-heading error, $\mathbb{I}_{\mathit{goal}}$ and $\mathbb{I}_{\mathit{col}}$ indicate goal reaching and collision, respectively, $\mathcal{P}_{\mathit{obs}}$ penalizes small obstacle distances, and $\|a_t-a_{t-1}\|^2$ penalizes abrupt action changes to encourage smooth control.

The safety-regularization term is

\begin{equation}
r_{\mathit{safe}}
=
-
c_v \mathcal{P}_v(v_t)
-
c_{\mathit{acc}}\mathbb{I}_{\mathit{risk}}\|\dot{v}_t\|^2,
\label{eq:r_safe}
\end{equation}

where $\mathcal{P}_v(v_t)$ penalizes excessive speed, $\dot{v}_t$ denotes the estimated linear acceleration, and $\mathbb{I}_{\mathit{risk}}$ activates the acceleration penalty in high-speed or attitude-risk states. This term suppresses sustained acceleration and abrupt braking, which may induce large inertia-related pitching moments and require large corrective control actions, thereby increasing transient attitude instability. Linear velocity is provided by the simulator for reward computation and critic value estimation in the asymmetric actor--critic architecture. Only the actor is deployed, so no additional linear-velocity measurement is required.

\section{EXPERIMENTS}
\label{sec:experiments}

This section evaluates SDOC through gradient analysis, comparative and ablation experiments, and hardware tests. We first introduce the setup, compared methods, and metrics, then analyze objective competition across attitudes and joint-task performance, and finally validate zero-shot transfer, disturbance response, and dynamic obstacle avoidance on hardware.

\subsection{Experimental Setup, Compared Methods, and Metrics}
\label{subsec:experimental_setup}

\subsubsection{Experimental Setup}

All policies are trained in parallel NVIDIA Isaac Sim environments with randomized obstacle layouts at each difficulty level. Evaluation follows the task and termination conditions in Section~\ref{subsec:task_formulation}, assessing joint-task performance on a fixed obstacle map and randomized obstacle maps at the highest difficulty level. All methods use the same evaluation maps, initial-pose distributions, and maximum episode length; other basic settings remain consistent except for the factors under investigation.

Except for the DDPG variant used for cross-algorithm validation, all policies are based on proximal policy optimization (PPO)~\cite{schulman2017proximal} and long short-term memory (LSTM) networks~\cite{graves2012long}. The same trained SDOC policy model is used for simulation evaluation and hardware deployment. Disturbances are introduced during simulation training to improve robustness, and the simulation evaluations in the main comparison table are also conducted under disturbances. The deployed actor has a model size below 8 MB and runs onboard on a Raspberry Pi 5.

\begin{figure}[!t]
  \centering
  \includegraphics[width=\columnwidth]{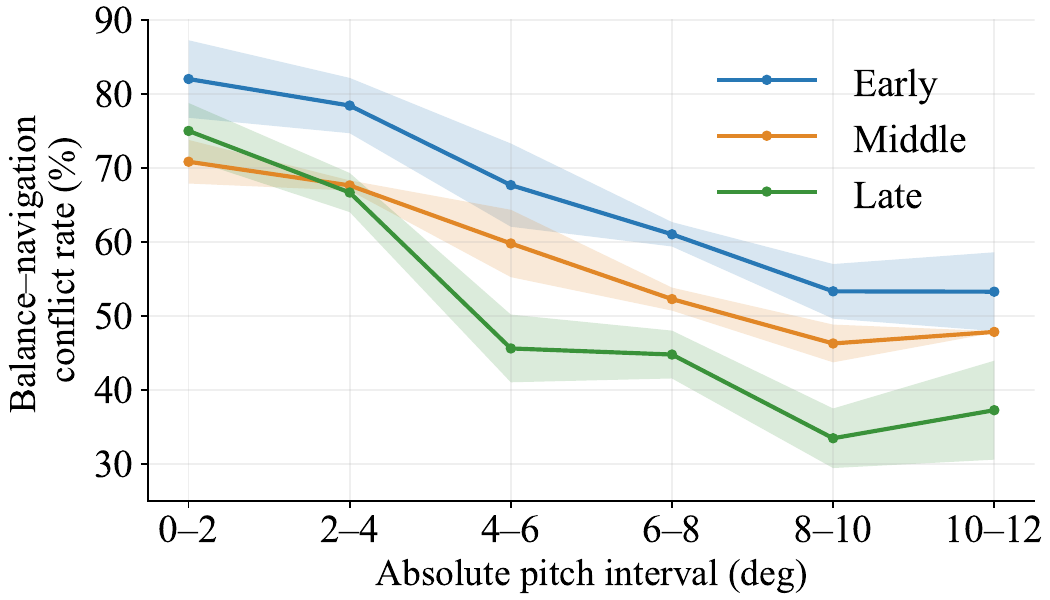}
  \caption{Gradient directional conflict rates between balance and navigation at fixed policy checkpoints. The three checkpoints are selected from the same training run. Policy parameters are frozen during gradient analysis, with no policy updates. The horizontal axis shows absolute pitch intervals, and the vertical axis shows the proportion of directional conflicts.}
  \label{fig:checkpoint_conflict_rate}
\end{figure}

\begin{figure}[!t]
  \centering
  \includegraphics[width=\columnwidth]{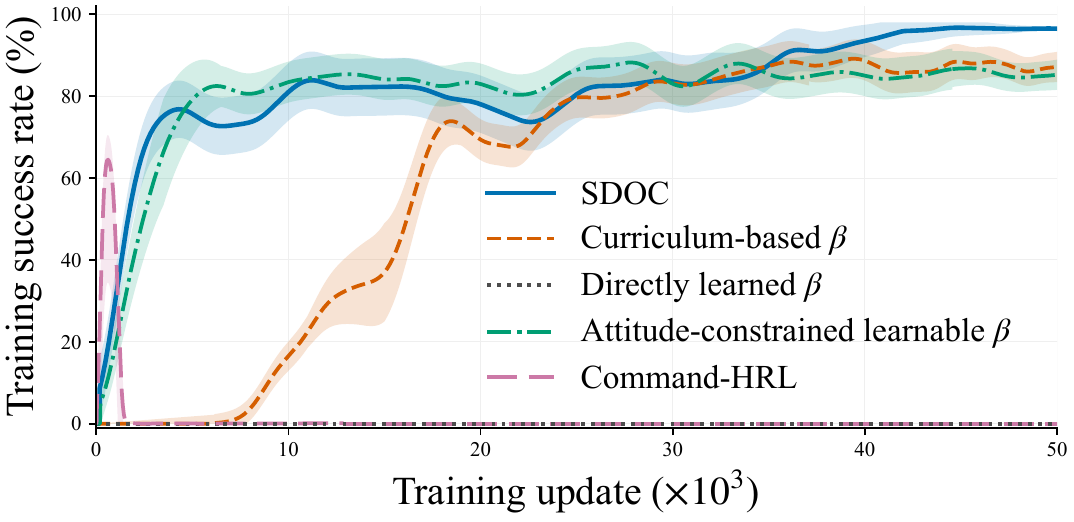}
  \caption{Training success rates of representative methods over training updates. Curves and shaded regions denote the mean and standard deviation across repeated training runs, respectively.}
  \label{fig:training_success_rate}
\end{figure}

\subsubsection{Compared Methods and Ablations}

Using SDOC, which determines reward weights directly from pitch, as the reference, we organize comparisons and ablations around reward coordination, training and control structures, and safety regularization. A DDPG variant additionally provides cross-algorithm validation.

\textbf{Reward coordination.} Fixed weighting uses a constant $\beta$ throughout training to examine a fixed objective trade-off. The other weighting schemes are described below.

\textbf{Curriculum-based weighting.} The balance reward weight $\beta$ starts at 1.0 and remains fixed within each curriculum stage. When navigation success stops improving and the adjustment criterion is met, $\beta$ is reduced to increase the relative navigation reward weight, guiding the transition from initial balance learning to joint balance and navigation learning.

\textbf{Directly learned weighting.} An additional actor output is clipped to a predefined range and used directly to weight the balance and navigation rewards, examining whether the policy can learn the objective trade-off autonomously. This output is not executed as a motor command.

\textbf{Attitude-constrained learnable weighting.} In addition to learning reward weights through the actor, this method imposes a weight upper bound that varies with pitch magnitude to examine coordination combining learned weights with an explicit attitude constraint.

We additionally include GCR-PPO~\cite{munn2026scalable} as a gradient-coordination baseline to compare coordination at the policy-optimization level with the proposed reward-level coordination on this task.

\textbf{Training and control structures.} Three alternatives examine stage-wise learning, hierarchical control, and the composition of independently trained policies.

\textbf{Stage-wise training.} The robot first learns to balance, after which navigation training continues from the trained balance model to examine whether the policy can acquire the joint task on top of its existing balancing ability.

\textbf{Command-based hierarchical reinforcement learning (Command-HRL).} A low-level controller is trained first, followed by a high-level command policy. The high-level policy generates commands from observations, and the low-level controller converts these commands into motor actions. This compares hierarchical control with a single end-to-end policy.


\textbf{Independent policy composition.} Balance and navigation policies are trained separately, and their action outputs are then blended with a fixed coefficient to examine whether independently learned control behaviors can accomplish the joint task through action composition.

\textbf{Safety-regularization ablation.} The safety-regularization term is removed from curriculum-based weighting while the other reward terms and termination conditions are retained. Comparison with its full-reward configuration examines the effect of safety regularization on navigation performance and attitude stability.

\textbf{Cross-algorithm validation.} The same state-dependent weighting map is used with DDPG~\cite{DBLP:journals/corr/LillicrapHPHETS15}, denoted by SDOC (DDPG), to examine the reward coordination scheme with another learning algorithm.

\begin{table*}[!t]
  \centering
  \caption{Comparison and ablation results on the fixed obstacle map and the hardest random obstacle map.}
  \label{tab:strategy_comparison}
  \scriptsize
  \normalfont
  \setlength{\tabcolsep}{0.5pt}
  \renewcommand{\arraystretch}{1.10}
  \begin{tabularx}{\textwidth}{@{}>{\raggedright\arraybackslash}p{0.225\textwidth}*{8}{>{\centering\arraybackslash}X}@{}}
    \toprule
    \multirow{2}{*}{Configuration}
    & \multicolumn{4}{c}{Fixed obstacle map}
    & \multicolumn{4}{c}{Hardest random obstacle map} \\
    \cmidrule(lr){2-5}\cmidrule(lr){6-9}
    & SR $\uparrow$ & AvgTime $\downarrow$ (s) & CR $\downarrow$ & $\overline{\theta}_{\mathit{max}}\downarrow$ (rad)
    & SR $\uparrow$ & AvgTime $\downarrow$ (s) & CR $\downarrow$ & $\overline{\theta}_{\mathit{max}}\downarrow$ (rad) \\
    \midrule
    \multicolumn{9}{l}{\textit{(a) Objective-coordination methods}} \\
    \midrule
    Fixed weight
    & \multicolumn{3}{c}{\textit{Fail (Fall)}} & --
    & \multicolumn{3}{c}{\textit{Fail (Fall)}} & -- \\
    Curriculum-based $\beta$
    & \multicolumn{3}{c}{\textit{Fail (Timeout)}} & $0.30\pm0.24$
    & \multicolumn{3}{c}{\textit{Fail (Timeout)}} & $0.28\pm0.04$ \\
    Directly learned $\beta$
    & \multicolumn{3}{c}{\textit{Fail (Timeout)}} & $0.32\pm0.22$
    & \multicolumn{3}{c}{\textit{Fail (Timeout)}} & $0.26\pm0.13$ \\
    Attitude-constrained learnable $\beta$
    & 57\% & $145.71\pm53.34$ & 1\% & $0.26\pm0.04$
    & 74\% & $112.06\pm31.85$ & 3\% & $0.24\pm0.13$ \\
    GCR-PPO
    & \multicolumn{3}{c}{\textit{Fail (Timeout)}} & $0.26\pm0.13$
    & \multicolumn{3}{c}{\textit{Fail (Timeout)}} & $0.29\pm0.14$ \\
    \textbf{SDOC (Ours)}
    & \textbf{98\%} & $\mathbf{21.20}\boldsymbol{\pm}\mathbf{2.50}$ & \textbf{1\%} & $\mathbf{0.14}\boldsymbol{\pm}\mathbf{0.02}$
    & \textbf{95\%} & $\mathbf{32.53}\boldsymbol{\pm}\mathbf{2.81}$ & \textbf{2\%} & $\mathbf{0.14}\boldsymbol{\pm}\mathbf{0.03}$ \\
    \midrule
    \multicolumn{9}{l}{\textit{(b) Training and control structures}} \\
    \midrule
    Stage-wise training
    & \multicolumn{3}{c}{\textit{Fail (Fall)}} & --
    & \multicolumn{3}{c}{\textit{Fail (Fall)}} & -- \\
    Command-HRL
    & 72\% & $15.70\pm4.21$ & 3\% & $0.22\pm0.04$
    & 2\% & $27.80\pm10.11$ & 0\% & $0.23\pm0.04$ \\
    Policy composition
    & 39\% & $86.07\pm4.73$ & 0\% & $0.39\pm0.02$
    & \multicolumn{3}{c}{\textit{Fail (Timeout)}} & $0.27\pm0.08$ \\
    \midrule
    \multicolumn{9}{l}{\textit{(c) Ablation and supplementary validation}} \\
    \midrule
    Curriculum-based $\beta$ (without safety regularization)
    & 50\% & $26.43\pm6.08$ & 3\% & $0.46\pm0.08$
    & 25\% & $62.22\pm53.96$ & 0\% & $0.51\pm0.12$ \\
    SDOC (DDPG)\textsuperscript{$\dagger$}
    & 92\% & $37.62\pm25.21$ & 2\% & $0.14\pm0.04$
    & 70\% & $59.80\pm21.94$ & 3\% & $0.59\pm0.52$ \\
    \bottomrule
    \end{tabularx}
    \par\vspace{2pt}
    \noindent
    \begin{minipage}{0.99\textwidth}
  \footnotesize
  \normalfont
  \textit{Notes.} Continuous-valued entries are mean $\pm$ standard deviation across independent evaluation runs; SR and CR are percentages. Metrics follow Section~\ref{subsec:experimental_setup}. \textit{Fail (Timeout)} and \textit{Fail (Fall)} denote all-timeout and all-fall evaluations, respectively; -- indicates pitch statistics are unreported due to all-fall evaluations. Configurations use PPO and the best-performing training checkpoint, except \textsuperscript{$\dagger$} DDPG, which uses the same $\beta(s)$ for cross-algorithm validation.
\end{minipage}

\end{table*}

\subsubsection{Evaluation Metrics}

We evaluate task completion, execution efficiency, attitude stability, and collisions. Let $\mathcal E_{\mathrm{succ}}$ denote the set of successful episodes, and let $N_{\mathrm{succ}}$ and $N_{\mathrm{eval}}$ denote the numbers of successful and total evaluation episodes, respectively.

Success Rate (SR) is the proportion of episodes that successfully reach the goal, $\mathrm{SR}=N_{\mathrm{succ}}/N_{\mathrm{eval}}$.

Average Completion Time (AvgTime) is computed over successful episodes only:

\begin{equation}
\mathrm{AvgTime}=\frac{1}{N_{\mathrm{succ}}}\sum_{i\in\mathcal E_{\mathrm{succ}}}T_i,
\label{eq:avg_time}
\end{equation}

where $T_i$ is the episode completion time. This metric is undefined when no episode succeeds.

The mean episode-wise maximum pitch magnitude measures attitude excursions during the task:

\begin{equation}
\overline\theta_{\mathrm{max}}=\frac{1}{N_{\mathrm{eval}}}\sum_{i=1}^{N_{\mathrm{eval}}}\max_{1\leq t\leq H_i}|\theta_{i,t}|,
\label{eq:mean_max_pitch}
\end{equation}

where $H_i$ is the episode length and $\theta_{i,t}$ is the pitch angle at step $t$ of episode $i$.

Collision termination rate (CR) is defined as $\mathrm{CR}=N_{\mathrm{col}}/N_{\mathrm{eval}}$, where $N_{\mathrm{col}}$ is the number of episodes ending upon satisfaction of the collision termination condition.

The metrics are first calculated within each evaluation run, and their means and standard deviations are then computed across independent evaluation runs.

\subsection{Gradient Conflict between Balance and Navigation Objectives}
\label{subsec:gradient_conflict}

To examine whether directional competition between balance and navigation varies with robot attitude, we select early, middle, and late policy checkpoints from the same training run: the first saved model after learning to balance, a model with some navigation success that is still steadily improving, and a model near peak training navigation success. At each checkpoint, we freeze policy parameters and collect data through repeated evaluations. Using the same sampled data, we compute the objective-specific policy gradients $g_{\mathrm{bal}}$ and $g_{\mathrm{nav}}$ without policy updates, identifying directional conflict by a negative inner product:

\begin{equation}
g_{\mathrm{bal}}^\top g_{\mathrm{nav}}<0.
\label{eq:gradient_conflict_condition}
\end{equation}
We partition the samples into attitude intervals according to absolute pitch and calculate the proportion of valid sample batches exhibiting directional conflict in each interval. Comparing pitch intervals within each checkpoint allows us to examine the relationship between objective competition and attitude under a fixed policy.

\begin{figure}[!t]
  \centering
  \includegraphics[width=\linewidth]{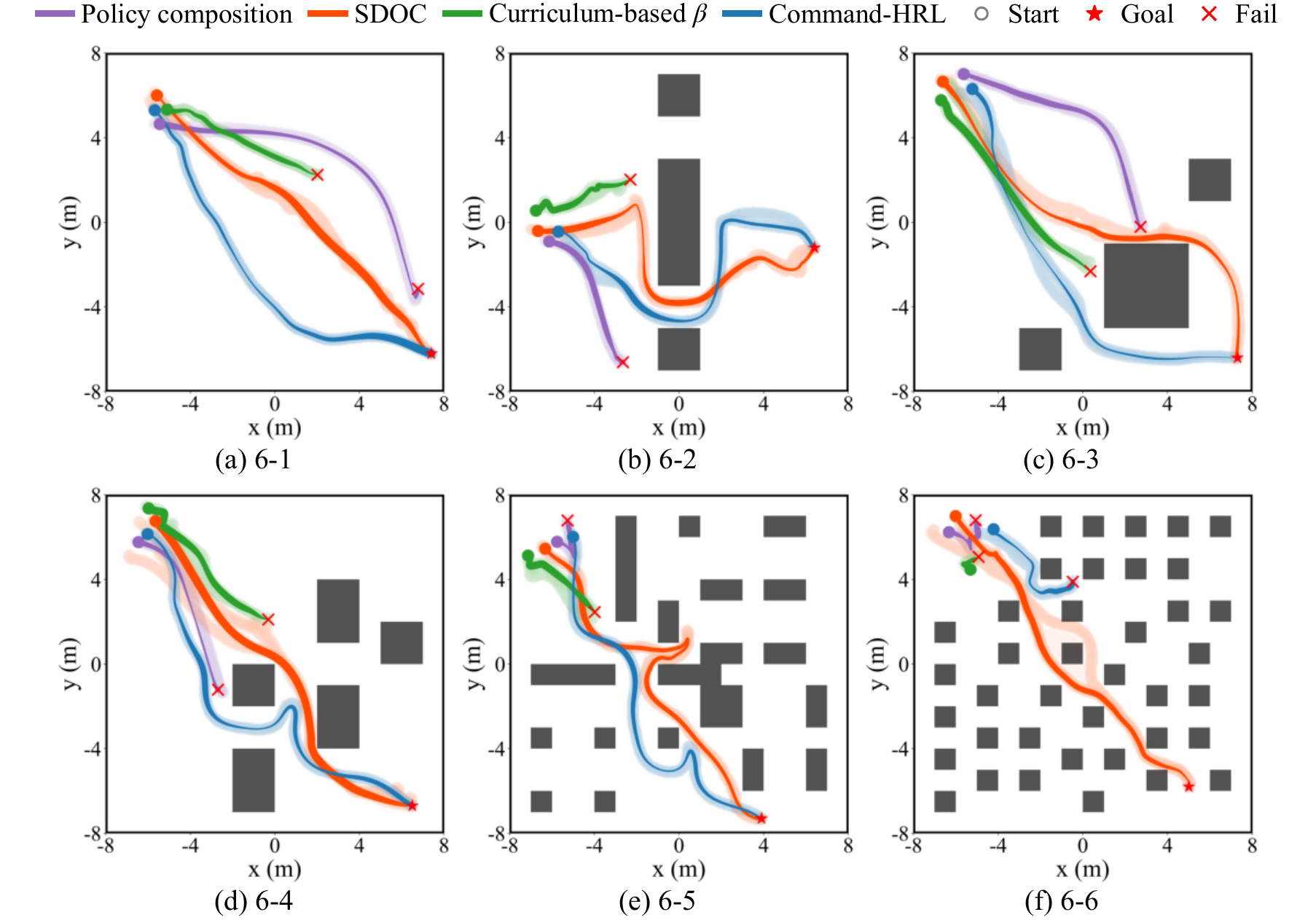}
  \caption{Two-dimensional trajectories of representative coordination strategies across six test maps.}
  \label{fig:traj_2x3_nogap}
\end{figure}

\begin{figure*}[!t]
  \centering
  \subfloat[]{\includegraphics[width=0.235\linewidth]{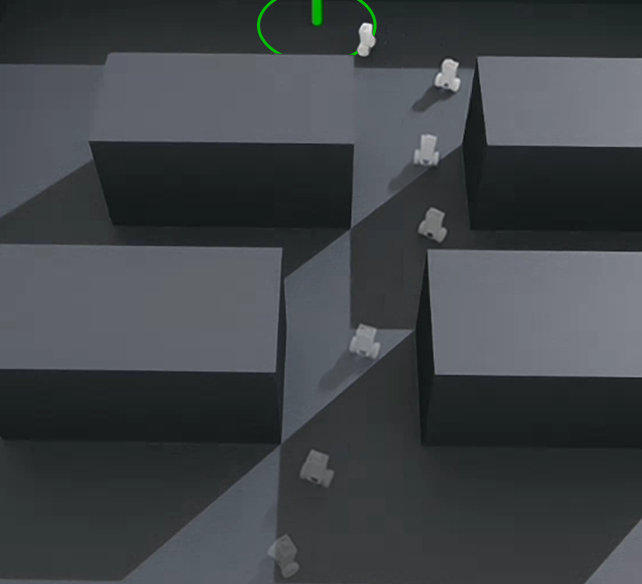}}%
  \subfloat[]{\includegraphics[width=0.235\linewidth]{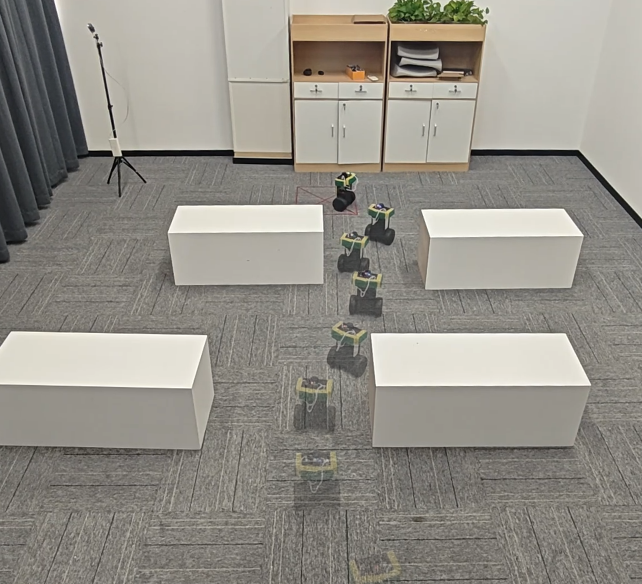}}%
  \subfloat[]{\includegraphics[width=0.235\linewidth]{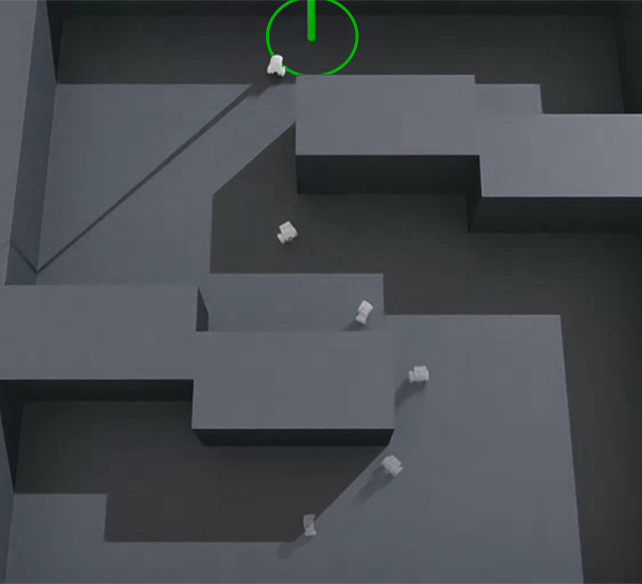}}%
  \subfloat[]{\includegraphics[width=0.235\linewidth]{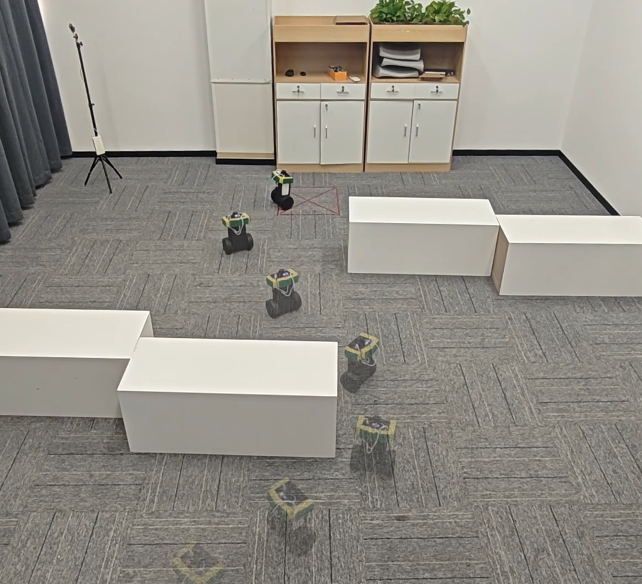}}

  \subfloat[]{\includegraphics[width=0.235\linewidth]{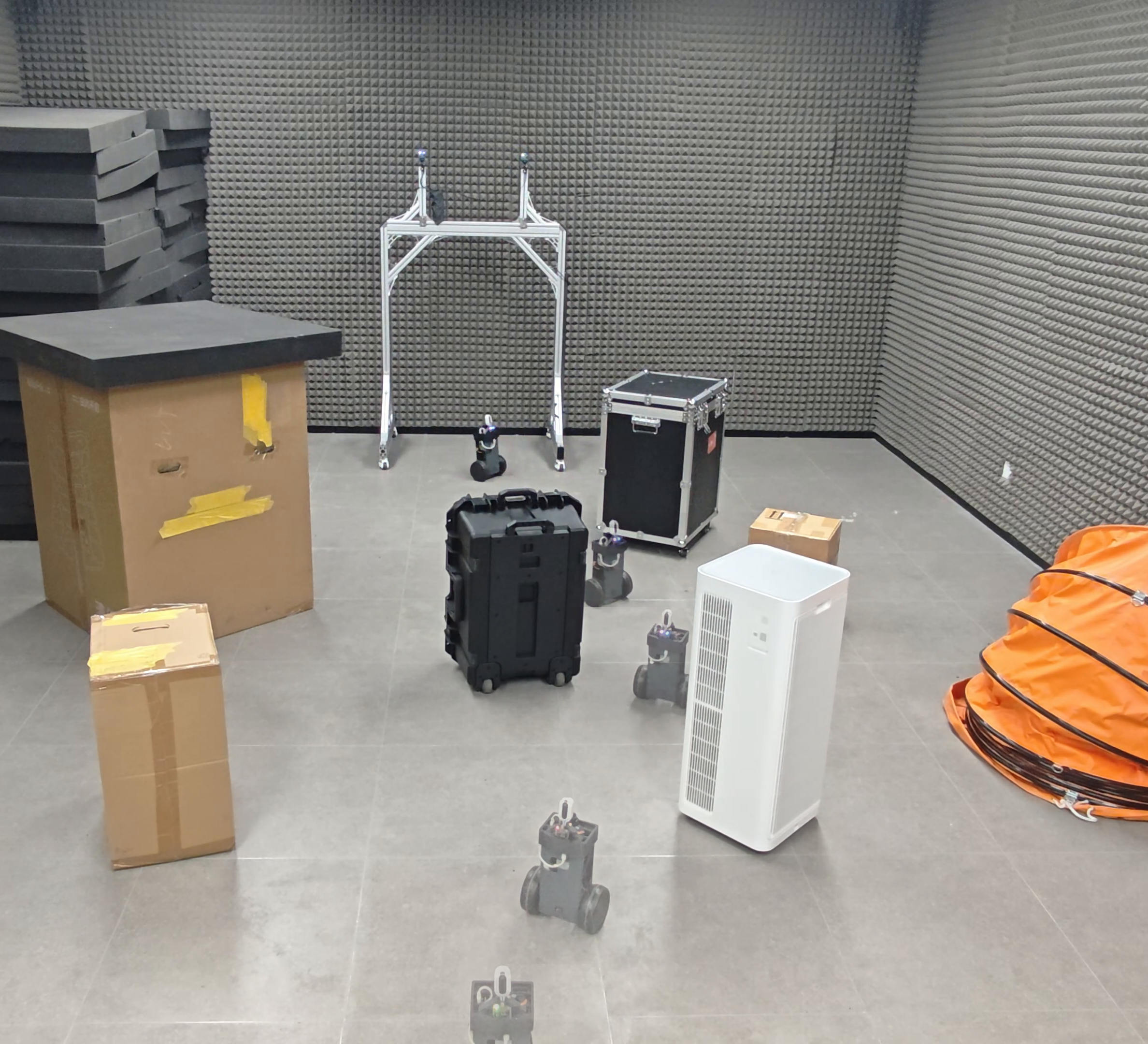}}%
  \subfloat[]{\includegraphics[width=0.235\linewidth]{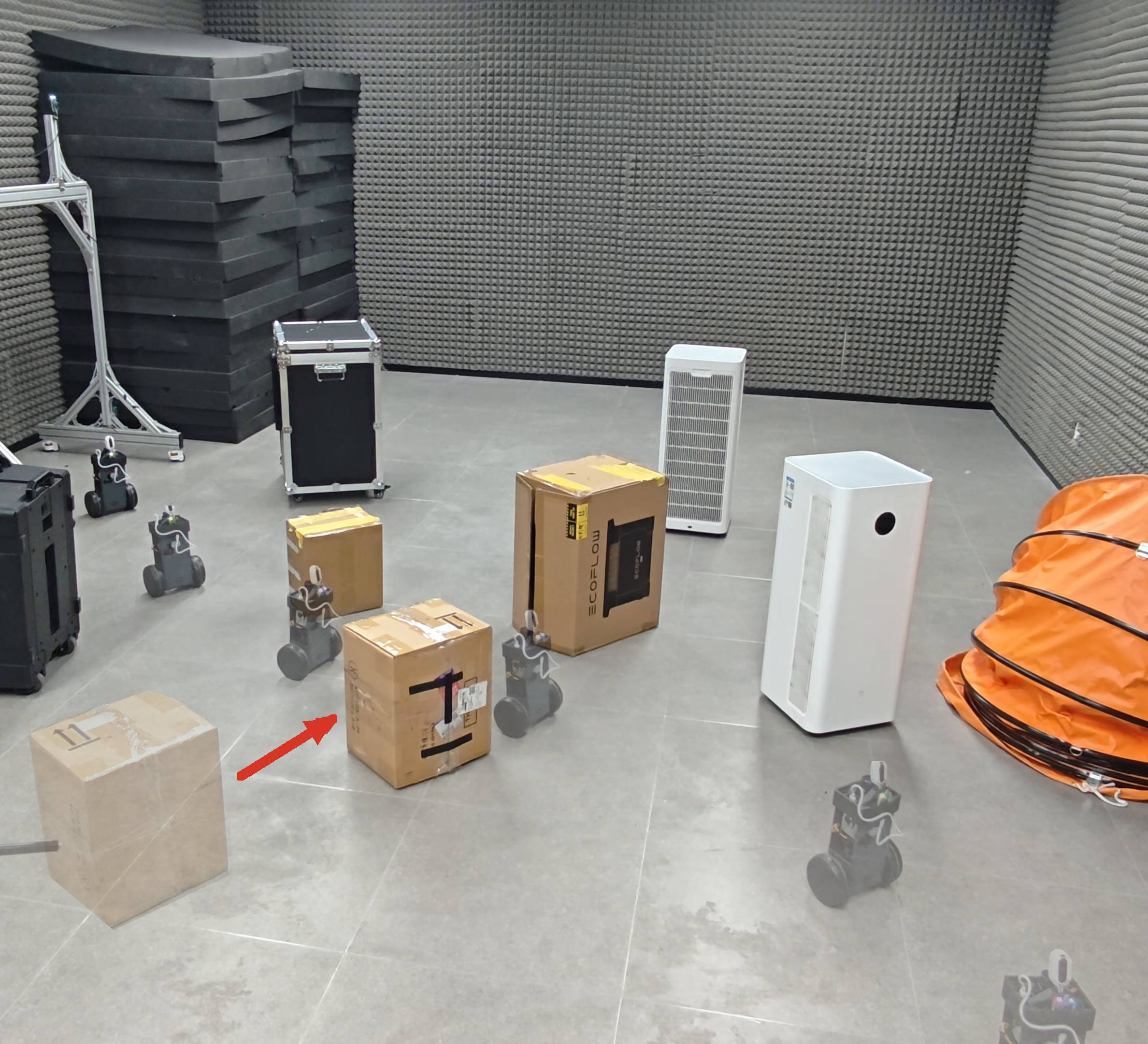}}%
  \subfloat[]{\includegraphics[width=0.235\linewidth]{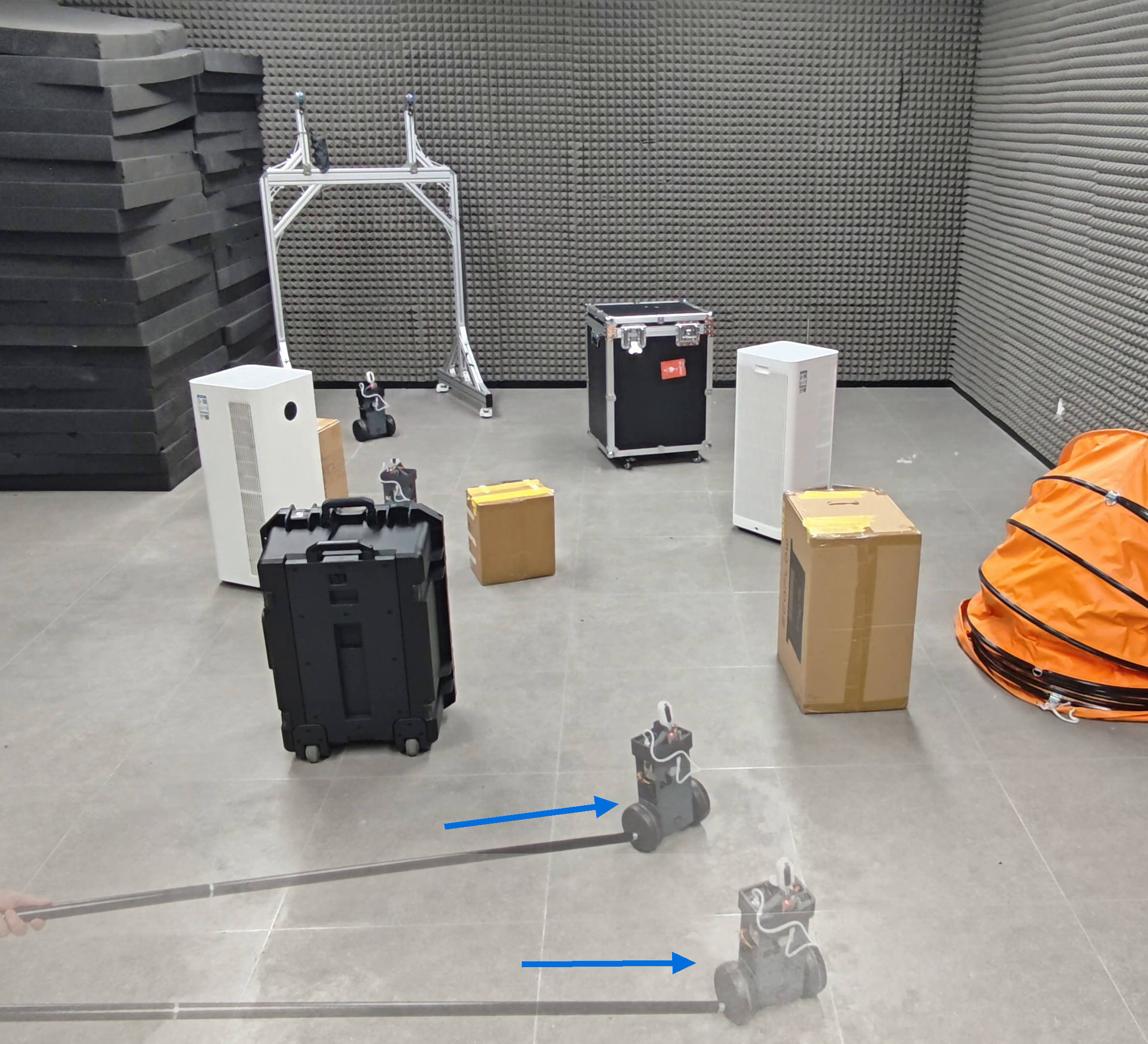}}%
  \subfloat[]{\includegraphics[width=0.235\linewidth]{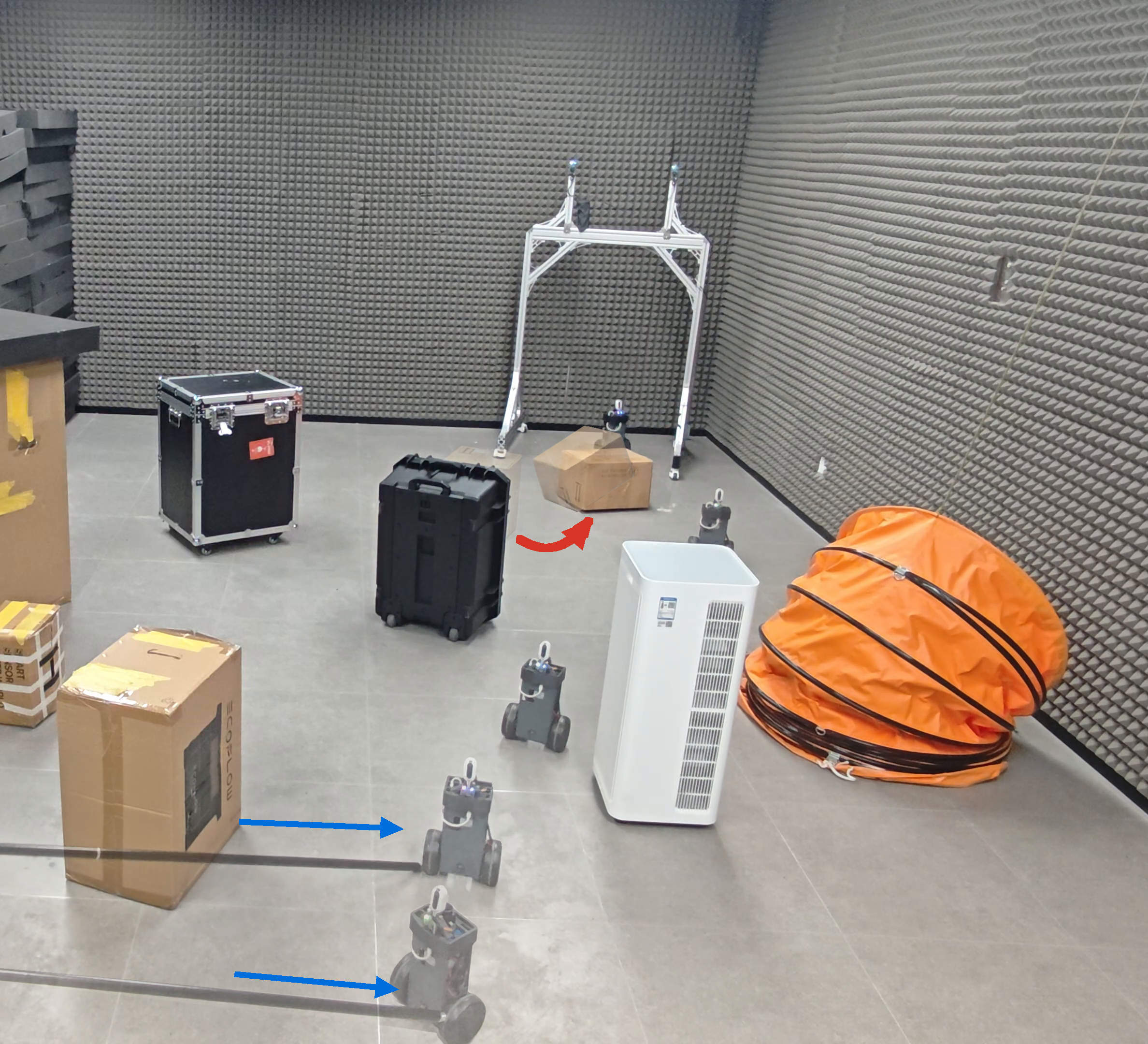}}

  \caption{Simulation-to-real-world comparison and hardware tests in unseen environments. (a), (c): simulation; (b), (d): hardware deployment in the corresponding layouts. (e)--(h): hardware trajectories in unseen layouts, including external disturbances in (g), (h) and moving obstacles in (f), (h). Red arrows indicate obstacle motion, and blue arrows indicate the directions of applied external disturbances.}
  \label{fig:sim2real_repro_2x4}
\end{figure*}

Fig.~\ref{fig:checkpoint_conflict_rate} presents conflict rates across pitch intervals for the three checkpoints. All three exhibit relatively high rates near upright and lower rates at larger pitch magnitudes. This indicates attitude-dependent directional competition and a need for coordination even near upright: maintaining stability and advancing navigation do not always favor aligned update directions. These observations motivate state-dependent reward coordination. Together with physical task requirements, they motivate greater relative emphasis on navigation near upright and increasing the balance reward weight with attitude deviation to promote recovery and sustain navigation.


\begin{figure}[!t]
  \centering

  \subfloat[Pitch]{\includegraphics[width=0.49\linewidth]{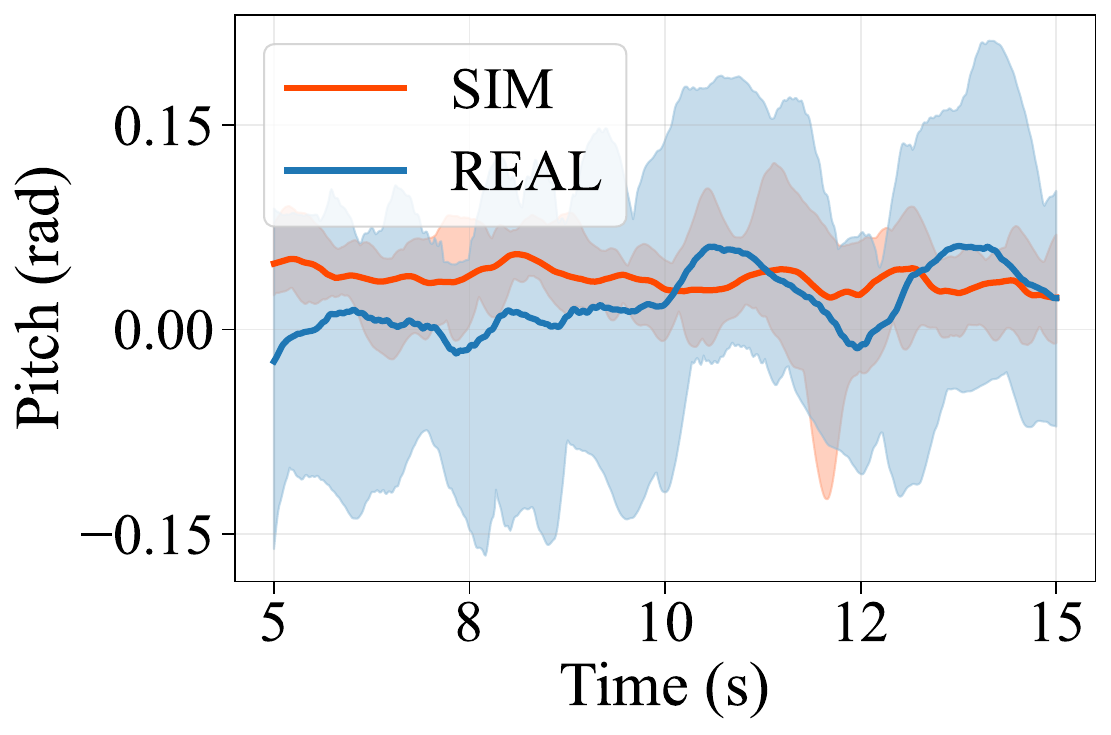}}%
  \subfloat[Pitch angular velocity]{\includegraphics[width=0.49\linewidth]{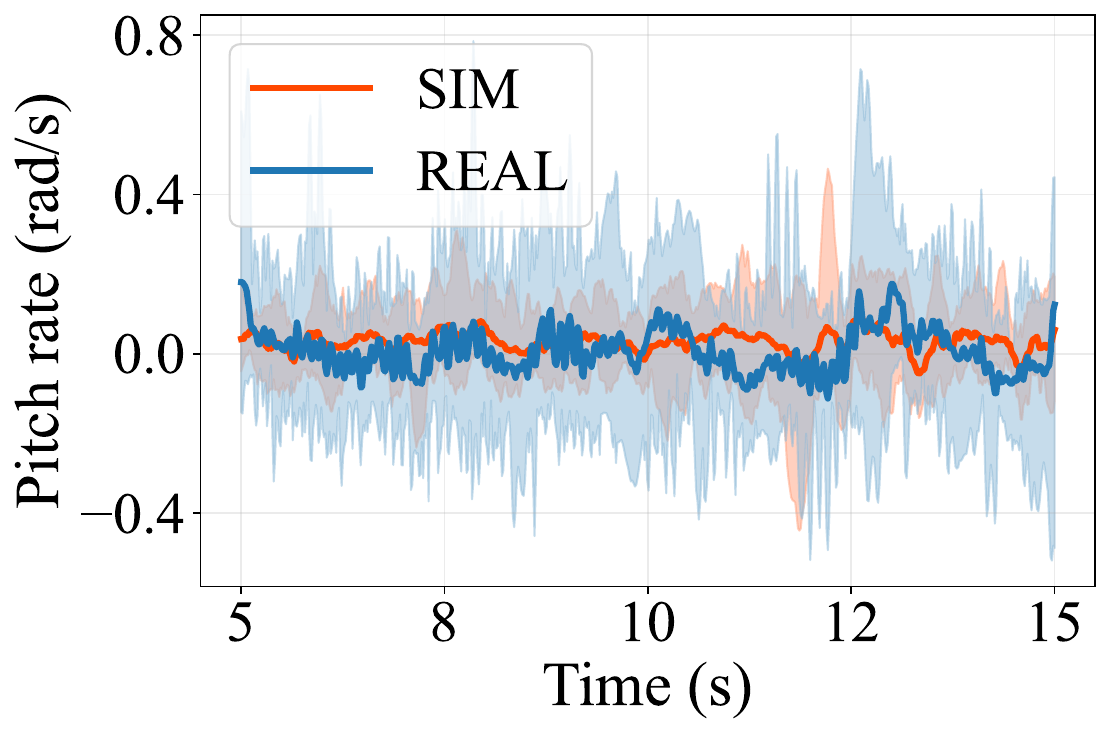}}

    \caption{Sim-to-real attitude signals of SDOC. Curves compare pitch and pitch angular velocity between simulation and hardware; shaded regions indicate variability across repeated hardware trials.}
  \label{fig:sim2real_timeseries}
\end{figure}

\subsection{Comparison and Ablation Results}
\label{subsec:comparison_ablation}

Fig.~\ref{fig:training_success_rate} compares training success rates. SDOC improves early, fluctuates, and reaches high success later; attitude-constrained weighting rises early and plateaus, while curriculum-based weighting improves later. Directly learned weighting remains near zero, and Command-HRL declines after an initial improvement.

Table~\ref{tab:strategy_comparison} compares performance on the fixed and hardest random obstacle maps, grouping methods into (a) objective coordination, (b) training and control structures, and (c) ablation and supplementary validation. Methods and metrics follow Section~\ref{subsec:experimental_setup}.

Table~\ref{tab:strategy_comparison}(a) shows that SDOC combines high success rates, short completion times, and stable attitude on both maps. Among objective-coordination methods, no tested fixed weight achieves both balance and navigation: the robot either stands until timeout or falls. Curriculum-based and directly learned weighting fail by timeout, while GCR-PPO makes no effective navigation progress. Attitude-constrained learnable weighting completes some tasks but remains less successful and efficient than SDOC.

Table~\ref{tab:strategy_comparison}(b) compares training and control structures. Stage-wise training fails by falling, while Command-HRL and policy composition show marked success-rate reductions on the hardest random map.

Table~\ref{tab:strategy_comparison}(c) examines safety regularization and cross-algorithm validation. Removing safety regularization allows curriculum-based weighting to complete some tasks but causes larger attitude excursions, reflecting instability risks associated with more aggressive navigation. SDOC (DDPG) further demonstrates the coordination mechanism's feasibility with another reinforcement learning algorithm.

Fig.~\ref{fig:traj_2x3_nogap} shows navigation trajectories of representative methods on six test maps, visualizing route continuity, rollout dispersion, navigation progress, and failure locations. Line thickness indicates pitch magnitude, with thicker trajectories representing larger attitude deviations.

\subsection{Zero-Shot Simulation-to-Real-World Validation}

Finally, we directly deploy the simulation-trained SDOC policy on the physical two-wheeled inverted-pendulum robot. Runtime processing includes only onboard observation construction, deterministic motor-command conversion, and necessary safety clipping. Fig.~\ref{fig:sim2real_repro_2x4}(a)--(d) compares simulation and hardware trajectories under two matched layouts, where the evaluation focuses on the consistency of obstacle-avoidance directions, turning behaviors, and goal-directed motion rather than exact geometric trajectory overlap.

Fig.~\ref{fig:sim2real_repro_2x4}(e)--(h) presents additional hardware trials in unseen obstacle layouts using the same deployed policy. Panel (e) shows navigation among static obstacles, panel (f) introduces a moving cardboard box, panel (g) shows external disturbances applied to the robot with a rod, and panel (h) combines rod-applied disturbances with a moving obstacle. The blue arrows indicate the directions of the applied disturbances, while the red arrows indicate obstacle motion. These image sequences provide qualitative evidence of continued balance and navigation under the tested layout changes, external disturbances, and obstacle motion.

\begin{table}[!t]
  \centering
  \caption{Hardware evaluation results of SDOC.}
  \label{tab:real_world}
  \small
  \setlength{\tabcolsep}{3.5pt}
  \renewcommand{\arraystretch}{1.08}
  \begin{tabular*}{\columnwidth}{@{\extracolsep{\fill}}cccc}
    \toprule
    SR & AvgTime (s) & $\mathrm{mean}(|\theta|)$ (rad) & $d_{\min}$ (m) \\
    \midrule
    $95\%\pm5\%$ & $17.09\pm9.04$ & $0.05\pm0.03$ & $0.16\pm0.02$ \\
    \bottomrule
  \end{tabular*}
\end{table}

Table~\ref{tab:real_world} summarizes task completion, attitude deviation, and minimum LiDAR range during hardware evaluation. AvgTime is calculated from successful trials. Mean absolute pitch is computed from onboard measurements and characterizes average attitude deviation, unlike the episode-wise maximum pitch used in the simulation comparison. The minimum LiDAR range, $d_{\min}$, is the smallest LiDAR range over each complete evaluation trial; the table reports the mean and standard deviation of these trial-wise minima.

Fig.~\ref{fig:sim2real_timeseries} further compares the pitch and pitch-angular-velocity trajectories between simulation and hardware. The SDOC hardware signals remain bounded, with pitch and pitch-angular-velocity ranges consistent with simulation. Together with the high success rate and consistent obstacle avoidance under matched layouts, these results demonstrate zero-shot transfer of the learned balance--navigation policy without policy adaptation.

\section{DISCUSSION AND CONCLUSION}

This work investigates end-to-end integrated attitude stabilization and obstacle-avoiding navigation for a two-wheeled inverted-pendulum robot. Since both objectives share the same wheel-torque commands, their relative importance varies with the attitude state. The proposed SDOC method coordinates balance and navigation objectives during training by adapting their scalarization according to the robot state. This coordination only affects reward construction, while the deployed system remains a single actor without additional control hierarchies, online scalarization, or auxiliary stabilizing controllers.

Gradient analysis at fixed policy checkpoints reveals attitude-dependent directional competition between the balance and navigation objectives. Comparative experiments show that SDOC combines high task success rates, navigation efficiency, and small attitude excursions on both obstacle maps. The DDPG experiments further demonstrate that the same coordination mechanism can be used with another training algorithm. Hardware experiments validate zero-shot transfer of the learned policy for joint balance and obstacle-avoiding navigation.


SDOC uses pitch magnitude to construct a compact state-to-weight mapping. Future work will extend this mapping to incorporate pitch angular velocity, obstacle proximity, actuator limits, and recoverability margins, and evaluate the method across additional robot platforms and deployment conditions.

\bibliographystyle{IEEEtran}
\bibliography{references}

\end{document}